\documentclass{article}

\usepackage[preprint]{neurips_2024}

\usepackage[utf8]{inputenc} 
\usepackage[T1]{fontenc}    
\usepackage{hyperref}       
\usepackage{url}            
\usepackage{booktabs}       
\usepackage{amsfonts}       
\usepackage{nicefrac}       
\usepackage{microtype}      
\usepackage{xcolor}         
\usepackage{makecell}
\usepackage{amsmath}
\usepackage{graphicx} 
\usepackage{booktabs}
\usepackage{multirow}
\usepackage{natbib}
\usepackage{tikz}
\usepackage{subcaption}
\usetikzlibrary{positioning}
\title{A Training Framework for Optimal and Stable Training of Polynomial Neural Networks}
\author{%
  Forsad Al Hossain \\
  Manning College of Information and Computer Science\\
  University of Massachusetts Amherst\\
  Amherst, MA 01002, USA \\
  \texttt{forsadalhossain@gmail.com} \\
  \And
  Tauhidur Rahman \\
  Hal\i c\i o\u{g}lu Data Science Institute\\
  University of California, San Diego \\
La Jolla, CA 92093, USA \\
  \texttt{trahman@ucsd.edu} \\
}

\begin{document}

\maketitle
\begin{abstract}
By replacing standard non-linearities with polynomial activations, Polynomial Neural Networks (PNNs) are pivotal for applications such as privacy-preserving inference via Homomorphic Encryption (HE). However, training PNNs effectively presents a significant challenge: low-degree polynomials can limit model expressivity, while higher-degree polynomials, crucial for capturing complex functions, often suffer from numerical instability and gradient explosion. We introduce a robust and versatile training framework featuring two synergistic innovations: 1) a novel Boundary Loss that exponentially penalizes activation inputs outside a predefined stable range, and 2) Selective Gradient Clipping that effectively tames gradient magnitudes while preserving essential Batch Normalization statistics. We demonstrate our framework's broad efficacy by training PNNs within deep architectures composed of HE-compatible layers (e.g., linear layers, average pooling, batch normalization, as used in ResNet variants) across diverse image, audio, and human activity recognition datasets. These models consistently achieve high accuracy with low-degree polynomial activations (such as degree 2) and, critically, exhibit stable training and strong performance with polynomial degrees up to 22, where standard methods typically fail or suffer severe degradation. Furthermore, the performance of these PNNs achieves a remarkable parity, closely approaching that of their original ReLU-based counterparts. Extensive ablation studies validate the contributions of our techniques and guide hyperparameter selection. We confirm the HE-compatibility of the trained models, advancing the practical deployment of accurate, stable, and secure deep learning inference.
\end{abstract}

\section{Introduction}
\label{sec:introduction}
In the modern era, cloud computing has emerged as a crucial infrastructure for deep learning, offering unparalleled scalability and flexibility by offloading the heavy computation of large deep learning models to the cloud. However, this reliance on external infrastructure intensifies concerns regarding data privacy and security, particularly when sensitive information is processed. Traditional methods often struggle to adequately protect data against increasingly sophisticated threats while complying with regulations like HIPAA \cite{HIPPA} and GDPR \cite{GDPR}, emphasizing the need for innovative solutions that enable secure computation on sensitive data.

A promising approach to address these privacy challenges in deep learning inference is the use of Homomorphic Encryption (HE), which allows computation directly on encrypted data. Polynomial Neural Networks (PNNs) have garnered significant attention in this context because their core operations (additions and multiplications) align well with the capabilities of current HE schemes \cite{cyrpto_net, transformer_poly, He_cnn, low_lat_he}. Unlike conventional networks using activations like Sigmoid or ReLU, which are computationally expensive or require approximations under HE, PNNs utilize polynomial activation functions, offering a more direct path to HE compatibility. Beyond their suitability for HE, PNNs possess other desirable properties. They can enhance model interpretability \cite{Fronk2022InterpretablePN, pi_net, meta_pol} and potentially improve generalizability \cite{pol_gen}, attributes valuable in domains requiring transparency and predictable behavior. While early theory suggested limitations \cite{universal_approx}, recent work indicates that deep PNNs can indeed act as universal approximator \cite{poly_universal_approx, poly_good}, further motivating their study. However, the potential of PNNs is hindered by significant training challenges, notably numerical instability and exploding gradients. This paper introduces a novel training framework designed specifically to overcome these obstacles and enable the stable and effective training of PNNs for HE applications. \\

\textbf{Our main contributions:}
We introduce a novel training framework for Polynomial Neural Networks (PNNs) that significantly enhances stability and performance which is essential for applications like Homomorphic Encryption (HE). Our primary contributions are:
\begin{itemize}
    \item \textbf{Innovative Training Techniques:} We propose two synergistic methods: 
    \begin{enumerate}
        \item A \textit{Boundary Loss} to exponentially penalize activation inputs outside a predefined stable range, preventing divergence.
        \item \textit{Selective Gradient Clipping} to manage large gradients while preserving Batch Normalization statistics by excluding its parameters.
    \end{enumerate}
    These enable stable training of PNNs with high-degree polynomials (e.g., up to degree 22) and performance comparable to ReLU counterparts, where standard methods often fail.

    \item \textbf{Demonstrated Generalizability and Performance:} Our framework's efficacy is validated across seven diverse datasets (image, audio, human activity recognition). Trained PNNs consistently achieve high accuracy, approaching or matching ReLU baselines, even with low-degree (e.g., degree 2) and also with high-degree polynomial activation replacements.

    \item \textbf{Comprehensive Empirical Validation:} Extensive ablation studies confirm the crucial synergy between our Boundary Loss and Selective Gradient Clipping for stability and accuracy, especially for higher-degree polynomials. We also offer practical hyperparameter tuning guidance.
\end{itemize}

In summary, we deliver a robust solution for training optimal and stable PNNs, facilitating their use in deep architectures and for privacy-preserving applications.

\section{Related Work}

\subsection{Polynomial Activation Approximation Strategies}
Various strategies exist for approximating non-linear activations with polynomials. Early work includes using Chebyshev polynomials for sigmoid approximation to derive ReLU derivatives \cite{poly_dl1}. Frameworks like Self-Learning Activation Functions (SLAF) \cite{slaf} enabled joint coefficient-weight optimization. Many approaches utilize fixed polynomials, either predetermined or learnable, to approximate activations (\cite{poly_dl1, poly_dl2, polynet_dist, cyrpto_net, poly_dl4, hermite_pol}). Other methods, such as those based on the Remez algorithm \cite{Remez_algo, Remez1, Remez2}, aim to minimize the maximum approximation error. Dynamic programming has been explored for layer-wise polynomial degree selection \cite{dyanamic_prog_pol_deg}, and learnable polynomial coefficients have also been shown to be effective \cite{trainable_coeff}.

\subsection{Challenges in Polynomial Neural Network Training}
Training Polynomial Neural Networks (PNNs) is notably challenging due to numerical instability with high-degree approximations \cite{poly_bad2} and gradient explosion \cite{poly_bad}. These difficulties are exacerbated when targeting Homomorphic Encryption (HE) applications, as HE schemes like CKKS \cite{CKKS} impose limitations on multiplicative depth, restricting PNN flexibility. While previous efforts have sought to address these issues, for instance by modifying input properties \cite{poly_bad, stable_input, scale_input_polynomial} or using client-assisted computations \cite{network_he, network_he2}, they often demonstrate stability only in narrow settings and lack broad empirical validation. Addressing these significant hurdles in stability and generalizability is crucial for unlocking the potential of PNNs, motivating our development of a robust training framework.
\section{Methodology}

\subsection{Model Architecture}
\label{sec:model_arch}

For homomorphic encryption (HE) compatibility, our neural network architectures exclusively used the following layers:

\begin{itemize}
    \item \textbf{Linear Layers}: Convolutional (for image/audio tasks) or fully connected (for MLPs).
    \item \textbf{Polynomial Activation (PNN)}:  Activations that maps a sample input with a polynomial function (i.e $f(x) = \sum_{i=0}^{n} a_{i}x^{i}$).
    \item \textbf{Average Pooling}: For downsampling in CNNs (e.g., replacing max-pooling in ResNet-18).
    \item \textbf{Batch Normalization}: Standard BatchNorm \cite{batch_norm}, fused into adjacent linear layers as affine transformations during inference for HE compatibility, preserving normalization.
    \item \textbf{Dropout}: For training regularization. During inference, dropout layers act as identity functions (effectively removed), posing no HE compatibility issues.
\end{itemize}

We excluded non-polynomial activations and other layers (e.g., Max-pool, Layer-Norm) not expressible by additions/multiplications, as they complicate HE implementation.
\begin{figure}[h]
\centering
\includegraphics[height=0.3\textheight, width=1.0\textwidth]{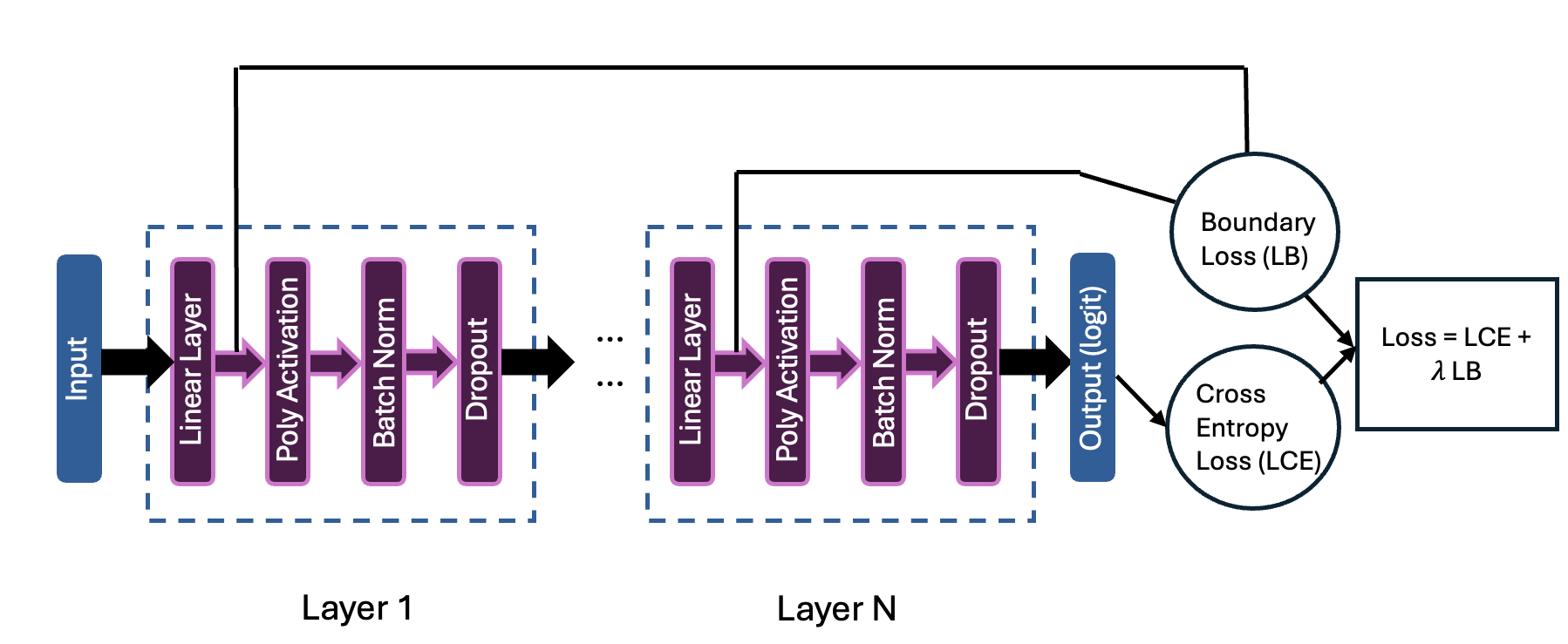}
\caption{Example of a neural network layer structure for our use cases. It only consists of layers which can be either dropped or converted into affine/polynomial functions}
\end{figure}

\subsubsection{Polynomial Activation Initialization}

For a target activation function $g(x)$ (e.g., ReLU), we fit polynomials by sampling $m$ points at uniform interval in the range $[-B, B]$ for the target approximation function. The polynomial coefficients $a_i$ are obtained using the least squares method, which determines the set of coefficients that minimizes the sum of squared differences between the polynomial approximation and the target function.




\begin{figure}[h]
\centering
\includegraphics[width=0.7\linewidth]{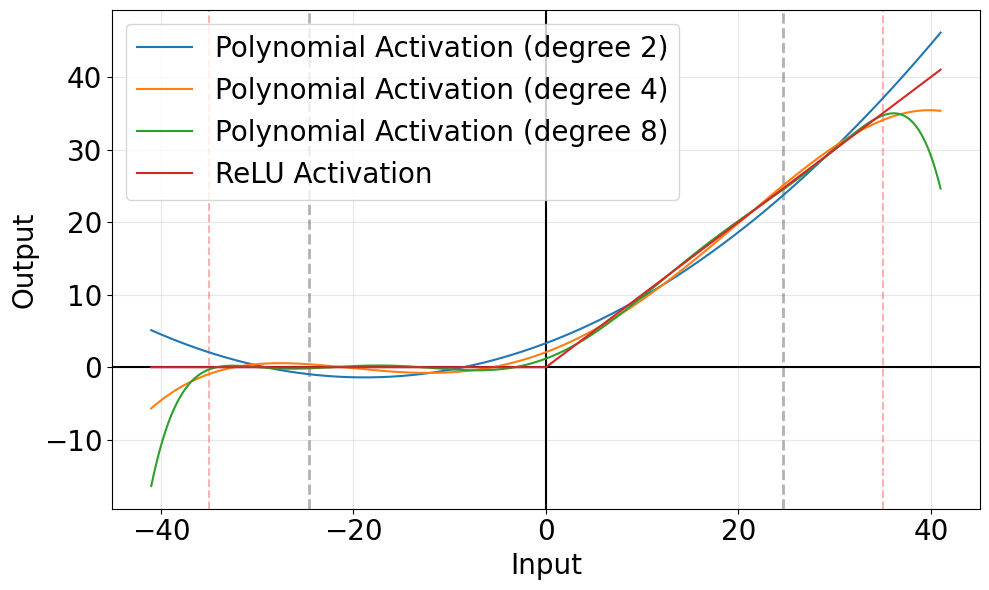}
\caption{Example of fitted polynomial activation functions for polynomial degrees of 2, 4, and 8. The activation function was fitted for the range $[-35, 35]$. The black dotted lines show the desired range of activation (if the value gets out of this bound we apply an exponential loss on the excess output)}
\end{figure}

\subsubsection{Boundary Constraint}

As our initial polynomial is only fitted for a particular range $[-B, B]$, and since the output of the polynomial is dominated by the highest power term, it is important that all inputs to activation functions remain inside the boundary at all times. To achieve this, we have introduced the following boundary loss term $L$ that exponentially penalizes activations that exceed the threshold $\alpha B$, where $\alpha \in (0,1]$.  The i'th layer boundary loss, is defined as:

\begin{equation}
L_{i} =  \frac{1}{|X_i|}\sum_{j=1}^{|X_{i}|}\left( e^{\max(|X_{i, j}| - \alpha B, 0)} - 1 \right)
\end{equation}

where $X_i$ is the input to the $i$-th layer activation. The introduction of the exponential term ensures steep penalties for significant boundary violations. With the introduction of this new loss, our composite loss function combines both classification and boundary losses from all activation input layers:

\begin{equation}
L_{\text{total}} = L_{\text{CE}} + \lambda \cdot \sum_{i=1}^{N} L_i
\end{equation}

where: $L_{\text{CE}}$ is the cross-entropy loss for the classification task, and N is the number of activation layers and $\lambda$ is weighting factor between these two losses
\subsection{Training Procedure}

\subsubsection{Selective Gradient Clipping}


 The exponential penalty of the boundary loss can cause large, destabilizing gradients. While norm-based gradient clipping (rescaling gradients above a norm threshold $c$) is standard, uniformly applying it to all parameters, including Batch Normalization (BatchNorm) layers, led to training instabilities in our experiments. This occurred because clipping BatchNorm's learnable parameters ($\gamma$ and $\beta$) hindered the adaptation of its running statistics, causing drift. To address this, we employ \textit{selective gradient clipping}, excluding BatchNorm parameters. For non-BatchNorm parameters $\theta_{\text{non-BN}}$, gradients are clipped as follows:

\begin{equation}
\nabla\theta_{\text{non-BN}}' = \frac{\nabla\theta_{\text{non-BN}}}{\max\left(1, \frac{\|\nabla \theta_{\text{non-BN}}\|_{2}}{c}\right)}
\label{eq:selective_clipping_update}
\end{equation}
Where $\|\cdot\|_{2}$ is the $L_{2}$ norm and $c$ is a predefined gradient clipping constant.
The gradients for BatchNorm parameters $\theta_{\text{BN}}$ remain unmodified ($\nabla\theta_{\text{BN}}' = \nabla\theta_{\text{BN}}$), allowing their statistics to update without the constraints imposed by clipping. This selective approach effectively stabilizes training by managing large gradients from the boundary loss without disrupting the crucial adaptive behavior of BatchNorm layers.

\subsubsection{Optimization}
\label{subsubsec:optimization}
We employed the AdamW optimizer \cite{AdamW} for model training, utilizing differential learning rates: a base rate, $\alpha_{base}$, for standard parameters, and a smaller rate, $\alpha_{poly} = 0.1 \times \alpha_{base}$, for the polynomial activation coefficients ($a_i$). For our case, we consistently set $\alpha_{base} = 0.001$. Additionally, an adaptive learning rate schedule was implemented, reducing the learning rate by a factor of $\gamma=0.1$ whenever validation performance plateaued for $p=5$ consecutive epochs. All models were trained for a fixed 30 epochs. However, for the CIFAR-100 \cite{cifar_100} dataset, training was extended to 120 epochs, with the learning rate starting at 0.001 and subsequently reduced by a factor of 10 at the 60th, 80th, and 100th epochs. As a general guideline, we set the B value in the range of 10-14 for degree 2 polynomials, 20-30 for degree 4 polynomials, and 35 when using degree 8 polynomials. We used $\alpha = 1$ for degree 2 polynomials, $\alpha = 0.75$ for degree 4 polynomials and $\alpha = 0.5$ for all higher degree polynomials for all of our training settings. The linear/convolutional layer weights were initialized with 'kaiming\_uniform' initialization \cite{kaiming_init} with 'linear' gain. The penalty $\lambda$ was consistently set to 1000, and dropout values between 0.0 and 0.2 were used, depending on the observed difference between validation loss and training loss (i.e. overfitting).




\section{Results}
\label{sec:results}


\newcommand{\polyrows}{3}
\begin{table}[hbtp]
\begin{tabular}{|c|c|c|c|c|}
\hline
Domain & Dataset & Pol degree &  PNN Accuracy & ReLU Accuracy \\ \hline
\multirow{12}{*}{\makecell[l]{Image \\ Classification}} & \multirow{\polyrows}{*}{MNIST} & 2 & 0.994  & \multirow{\polyrows}{*}{0.994}\\ \cline{3-4}
& & 4 & 0.994 & \\ \cline{3-4}
& & 8 & 0.994 & \\ \cline{2-5}

& \multirow{\polyrows}{*}{FashionMNIST}  & 2 & 0.92 & \multirow{\polyrows}{*}{0.93} \\ \cline{3-4}
& & 4 & 0.93 & \\ \cline{3-4}
& & 8 &  0.93 & \\ \cline{2-5}

& \multirow{\polyrows}{*}{CIFAR 10}  & 2 &  0.89 & \multirow{\polyrows}{*}{0.91} \\ \cline{3-4}
&  & 4 &  0.91 & \\ \cline{3-4}
& & 8 & 0.91 & \\ \cline{2-5}

& \multirow{\polyrows}{*}{CIFAR 100} & 2  & 0.63 & \multirow{\polyrows}{*}{0.64}  \\ \cline{3-4}
&  & 4 & 0.65 & \\ \cline{3-4}
&  & 8 & 0.66 & \\ \hline


\multirow{6}{*}{\makecell[l]{Human Activity \\Recognition}}  & \multirow{\polyrows}{*}{UCI-HAR} & 2 & 0.95 & \multirow{\polyrows}{*}{0.95} \\ \cline{3-4}
&  & 4 & 0.94 & \\ \cline{3-4}
&  & 8 & 0.94 & \\ \cline{2-5}

& \multirow{\polyrows}{*}{Capture 24}  & 2 & 0.74  &  \multirow{\polyrows}{*}{0.76}\\ \cline{3-4}
& & 4 & 0.75 & \\ \cline{3-4}
&  & 8 & 0.76 & \\ \hline


\multirow{3}{*}{\makecell[l]{Audio \\Classification}} & \multirow{\polyrows}{*}{Speech Commands}   & 2  & 0.96 & \multirow{\polyrows}{*}{0.97} \\ \cline{3-4}
&  & 4 & 0.96 & \\ \cline{3-4}
&  & 8 & 0.96 & \\ \hline


\end{tabular}
\newline \newline
\caption{Accuracy comparison between PNNs (trained with our method) and ReLU baselines across datasets}
\label{tab:datasets}
\end{table}


Polynomial neural networks (PNNs) were trained and evaluated using our proposed methodology across diverse datasets: MNIST \cite{mnist}, CIFAR-10 \cite{cifar10}, CIFAR-100 \cite{cifar_100}, FashionMNIST \cite{fashionmnist}, UCI-HAR \cite{ucihar}, Capture-24 \cite{capture24}, and Speech Commands \cite{speech_commands}. We utilized predefined training/test splits where available;except for Capture-24, a 10\% participant data subset formed the test set. Validation splits were predefined (Speech Commands) or derived from training data. Image and audio tasks employed a ResNet-18 architecture, while human activity recognition used a 7-layer MLP.
\section{Ablation studies}
\label{sec:ablation}
\subsection{Ablation study for different polynomial degree}

\begin{figure}
    \centering
    \includegraphics[width=1\linewidth]{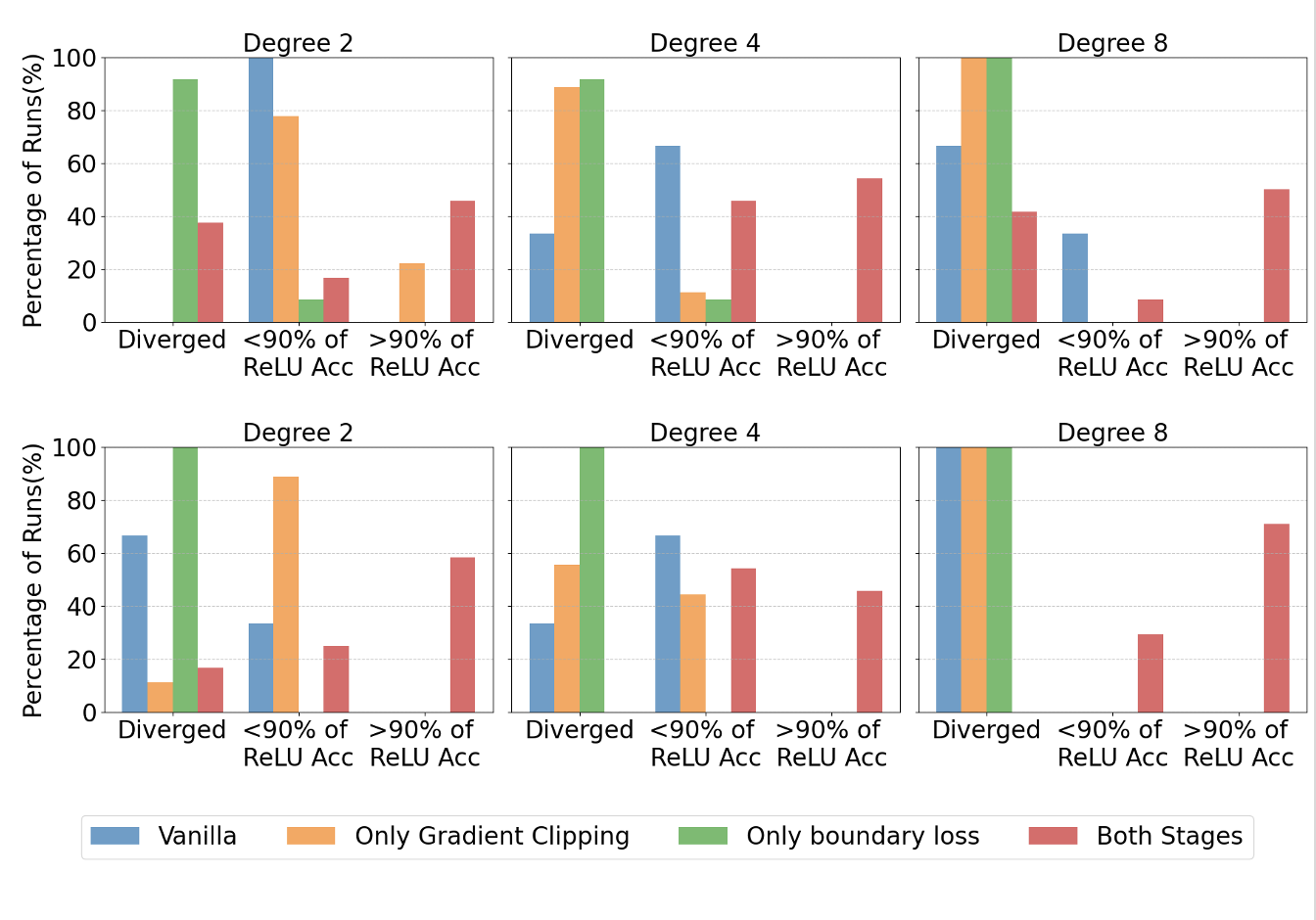}
    \caption{Evaluation of training stability and accuracy via ablation study on CIFAR-10 (top) and CIFAR-100 (bottom). The proposed training methodology demonstrates superior convergence rates and accuracy compared to baseline and partial implementations (boundary loss only, gradient clipping only) across varying polynomial degrees (2, 4, 8).}
    \label{fig:acc_distribution}
\end{figure}

Figure \ref{fig:acc_distribution} presents ablation studies on CIFAR-10 and CIFAR-100, exploring relevant hyperparameters for four training configurations. For PNN degrees 2, 4, and 8, $B$ values were (13, 20, 35) for CIFAR-10 and (14, 30, 35) for CIFAR-100, respectively. The configurations were:
\begin{itemize}
    \item \textbf{Baseline (No Modifications):} Excludes our Boundary Loss and Selective Gradient Clipping.
    \item \textbf{Boundary Loss Only:} Applies only our Boundary Loss (no gradient clipping).
    \item \textbf{Gradient Clipping Only:} Applies only standard gradient clipping (no Boundary Loss).
    \item \textbf{Proposed Method (Both Techniques):} Employs both our Boundary Loss and Selective Gradient Clipping.
\end{itemize}
Our proposed method, combining Boundary Loss and Selective Gradient Clipping, significantly outperformed the other configurations. Partial strategies (Baseline, Boundary Loss Only, or Gradient Clipping Only) often resulted in convergence failures or training instability, particularly for higher-degree polynomials, due to issues like unaddressed instability or exploding gradients. These findings (detailed in following subsections) highlight the crucial synergy of our combined techniques for stable and high-performing PNNs. We have also conducted statistical significance study between the success rates our methodology vs other settings and found that the difference in these success rates was found to be statistically highly significant (p < 0.01, two-sided Z-test), which implies that our methods are much more effective compared to the baseline.
The divergence occasionally observed in our method stems from the critical tuning of the B parameter and stochastic elements like dropout. For optimal training, B must be minimized (see Section \ref{sec:B_study}); however, setting B too low can lead to an exploding boundary loss, particularly when dropout increases random neuron firing during initialization. This highlights the balance needed in selecting B for performance versus stability against random initializations.

\subsubsection{Impact of Polynomial Degree within Ablation Study}
Our ablation studies, varying polynomial degrees (2, 4, 8), reveal distinct performance characteristics. Notably, even low-degree polynomials (e.g., degree 2) achieve strong accuracy, often exceeding 90\% of the ReLU baseline, when trained with our complete proposed method (Boundary Loss and Selective Gradient Clipping). While medium-degree polynomials (degree 4) can offer improvements, higher-degree polynomials (degree 8) generally yield the best results with our full framework, in some instances matching or surpassing ReLU baseline performance. Conversely, training with incomplete methods (baseline, or only Boundary Loss, or only Selective Gradient Clipping) leads to significantly poorer performance or complete training failure, particularly for higher-degree polynomials, underscoring the critical synergy of our proposed techniques.

\subsection{Effect of Polynomial Fitting Boundary ($B$) and Degree}
\label{sec:B_study}
\begin{figure}[h!]
\centering
\includegraphics[width=1.0\linewidth]{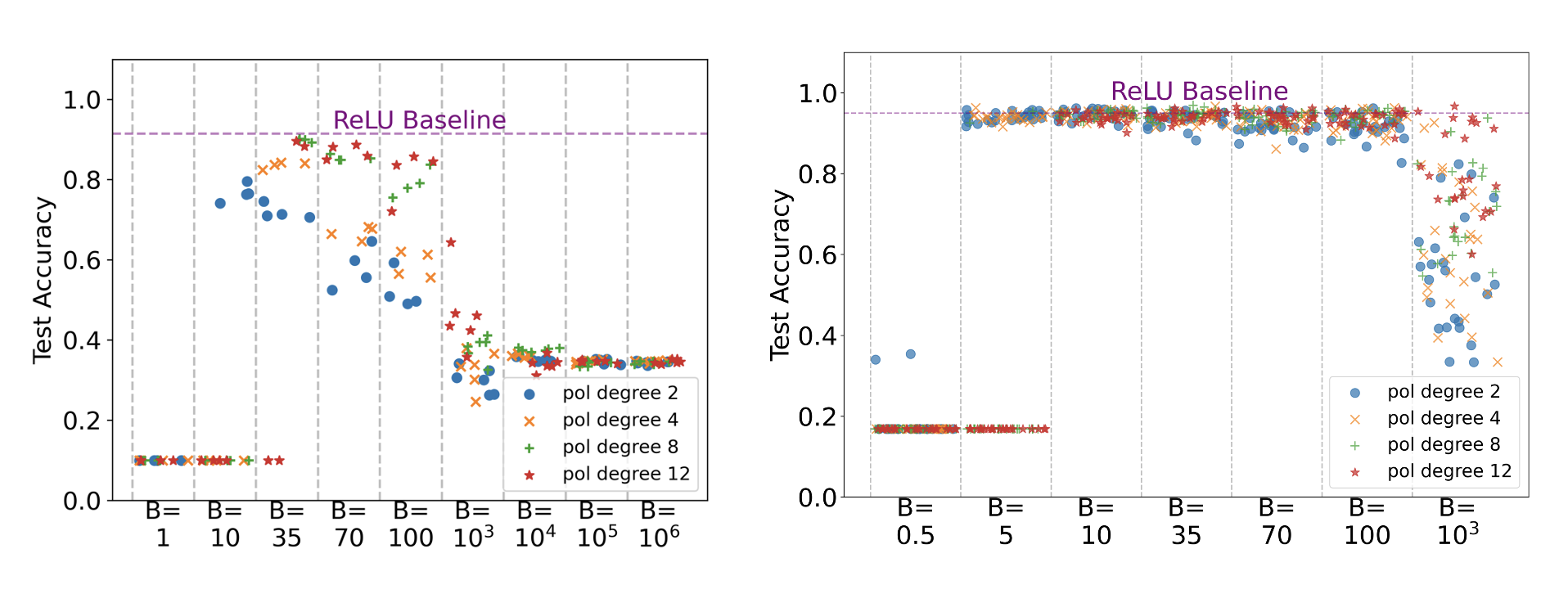} 
\caption{Impact of polynomial fitting boundary $B$ on test accuracy for CIFAR-10 (left) and UCI-HAR (right) datasets for polynomial degrees 2, 4, 8, and 12. Each dot represents a training run with varied parameters (e.g., dropout, gradient clipping, $\lambda$, fit samples). Small $B$ leads to instability, while excessively large $B$ degrades performance. Higher degrees generally perform better but require larger $B$ for stability. The dashed purple lines indicates ReLU baseline accuracy.}
\label{fig:B_plots}
\end{figure}

The polynomial fitting boundary $B$, defining the initial interval $[-B, B]$ for fitting activations, critically impacts training stability and model accuracy, as illustrated in Figure \ref{fig:B_plots} for CIFAR-10 and UCI-HAR. Key observations include:

\begin{itemize}
    \item \textbf{Instability at Low $B$:} Small $B$ values consistently destabilize training, as inputs often fall outside this narrow, reliable fitting range.
    \item \textbf{Stability vs. Performance Trade-off at High $B$:} Increasing $B$ enhances stability, but excessively large $B$ values can sharply degrade performance. This suggests that overly wide fitting ranges may poorly approximate the target non-linearity (e.g., ReLU) in critical input regions.
    \item \textbf{Interaction between $B$ and Polynomial Degree:} The optimal $B$ correlates with polynomial degree $d$. Lower-degree polynomials achieve optimal performance with smaller $B$, while higher degrees necessitate larger $B$ for stability. When appropriately stabilized, higher-degree polynomials tend to yield better accuracy (cf. Table \ref{tab:datasets}). 
    \item \textbf{Dataset and Network Dependency:} While the general trends for $B$ hold across datasets, the specific optimal $B$ values and achievable accuracies are dataset- and network-dependent.
\end{itemize}

In short, selecting an appropriate $B$ is critical and involves a dataset- and degree-specific trade-off, as noted in prior works \cite{hermite_pol, bound_b_train}. The best approach is typically to choose $B$ near the lower end of the stable training range for a given polynomial degree. This maximizes performance by preventing instability from overly narrow intervals and avoiding performance decline from excessively wide setting of $B$.

\subsubsection{Impact of Selective Gradient Clipping (Excluding Batch Norm Parameters)}

\begin{figure}[h!]
\centering
\includegraphics[width=1.0\linewidth]{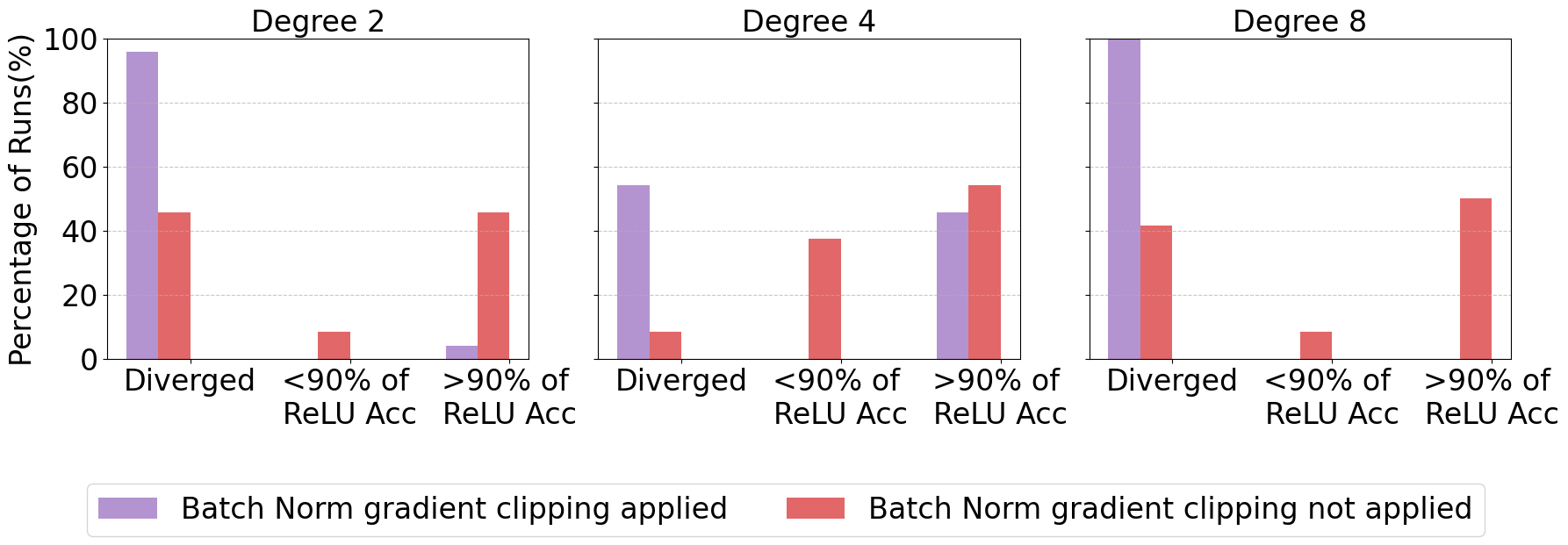}
\caption{Effect of including (purple bars) versus excluding (red bars; our proposed Selective Gradient Clipping) Batch Normalization parameters from gradient clipping for the CIFAR-10 dataset. Results are shown for PNNs with polynomial degrees 2, 4, and 8, fitted with B-values of 13, 20, and 35, respectively.}
\label{fig:bnorm_cifar10}
\end{figure}

Figure \ref{fig:bnorm_cifar10} underscores the significant benefits of our Selective Gradient Clipping strategy, which purposefully excludes Batch Normalization (BN) parameters from the gradient clipping process. When compared to an alternative approach that clips all parameters including those of BN (depicted by purple bars, versus the red bars representing our selective method on CIFAR-10), our technique demonstrates markedly superior training stability and accuracy across all tested polynomial degrees (2, 4, and 8). The inclusion of BN parameters in gradient clipping consistently leads to a higher incidence of training divergence and diminished performance. This detrimental effect is evident even with low-degree (degree 2) polynomials, where clipping BN parameters causes extensive training failures, a problem substantially alleviated by our selective approach. The importance of excluding BN parameters from clipping becomes critically amplified for higher-degree polynomials; for instance, with degree 8 PNNs, clipping BN parameters results in complete training collapse, whereas our selective method facilitates successful convergence and the achievement of high accuracy. These findings confirm that preserving the adaptive learning capability of Batch Normalization by omitting its parameters from gradient clipping is vital for the robust and effective training of PNNs, especially as the polynomial complexity increases.

We have also explored additional ablation studies to explore the sensitivity to other hyperparameters. For instance, the framework demonstrated robustness to the choice of the boundary loss penalty $\lambda$ (across values from 1 to 1000), variations in batch size(30-128), and different values of gradient clipping parameter (from 0.5 to 1.0), even with differnt activations such as SiLU, with performance remaining largely consistent in every case. Beyond the degrees explored in our main ablation studies, we also successfully trained polynomial degrees as high as 22 for the Speech Commands dataset. This further illustrates the potential of our framework to push the boundaries of PNN training complexity, although more extensive experimentation would be needed to fully characterize performance at such extreme degrees.
\section{Runtime and HE Implementation Details}

Our trained models demonstrate high Homomorphic Encryption (HE) compatibility. Using the CKKS scheme in OpenFHE \cite{OpenFHE} on an AMD EPYC 9654 CPU (64 threads), a 5-layer MLP on the UCI-HAR dataset achieved 94.7\% accuracy in encrypted inference, matching unencrypted performance. This utilized CKKS parameters for 128-bit security (ring dimension $N=2^{15}$, 40-bit scaling precision, multiplicative depth twice the layer count) and a quadratic polynomial activation. The HE inference time was 177.33 seconds per sample, which can be improved via OpenFHE optimization and input batching.

\section{Discussion}
\label{sec:discussion}
Training Polynomial Neural Networks (PNNs) for applications like Homomorphic Encryption (HE) is often hindered by numerical instability and gradient explosion. Our framework, integrating a novel \textbf{Boundary Loss} and \textbf{Selective Gradient Clipping}, directly addresses these critical challenges.

The synergistic efficacy of our approach is clear: ablation studies (Figure \ref{fig:acc_distribution}) demonstrate that both components are vital for stable convergence and high accuracy, particularly for higher-degree polynomials (e.g., degrees 4 and 8), outperforming baseline or partial methods. This framework's generalizability was shown by successfully training PNNs across seven diverse datasets. Our PNNs achieved accuracies comparable to standard ReLU networks (Table \ref{tab:datasets}), especially with degree 8 polynomials, highlighting their significant expressive power when properly stabilized. Furthermore, analyses of the polynomial fitting boundary $B$ (Figure \ref{fig:B_plots}) and the selective exclusion of BatchNorm parameters from gradient clipping (Figure \ref{fig:bnorm_cifar10}) confirmed their essential roles in optimizing performance and stability, offering practical hyperparameter guidance. By facilitating more robust and accurate PNNs, this work contributes to advancing privacy-preserving machine learning, enabling secure processing of sensitive data via HE across various domains.

\subsection{Limitations}
\label{subsec:limits}
Despite the demonstrated advancements, this work has several limitations.
\begin{itemize}
    \item \textbf{Inference Cost of High-Degree Polynomials:} While our framework enables stable training of PNNs with higher degrees, the inherent computational cost of evaluating these high-degree polynomials during inference remains a factor(because of multiplication depth constraints). This is particularly relevant for resource-constrained environments or when using HE schemes where polynomial degree significantly impacts performance or parameter choices.
    \item \textbf{PNN Transfer Learning:} This study focuses on training PNNs per dataset. The potential for transfer learning, such as pre-training PNNs on larger datasets and fine-tuning for specific tasks, especially in HE contexts, was not explored and remains an area for future work.
    \item \textbf{Empirical Focus:} The development and validation of our framework are primarily empirical. While results are strong across diverse datasets, a deeper theoretical analysis of the stability conditions or convergence guarantees provided by the combined effect of Boundary Loss and Selective Gradient Clipping could further solidify the approach.
\end{itemize}

\section{Conclusion}

In this paper, we presented a novel training framework designed to enable stable and effective training of high-degree Polynomial Neural Networks. By introducing a Boundary Loss to constrain activation inputs and employing Selective Gradient Clipping to manage gradient magnitudes without disrupting Batch Normalization, we successfully trained PNNs with degrees up to 8 across a diverse set of benchmarks, achieving performance comparable to ReLU baselines in several cases.

Our work facilitates the development of deep learning models suitable for privacy-preserving inference using Homomorphic Encryption. The demonstrated stability and generalizability provide a foundation for further exploration in this domain.

\bibliographystyle{plainnat}
\bibliography{cite.bib}



\end{document}